\documentclass[tinyml]{acmart}

\usepackage{makecell}
\usepackage{gensymb}

\AtBeginDocument{%
  \providecommand\BibTeX{{%
    \normalfont B\kern-0.5em{\scshape i\kern-0.25em b}\kern-0.8em\TeX}}}

\setcopyright{rightsretained}
\copyrightyear{2022}
\acmYear{2022}

\begin{document}

\title{Power-of-Two Quantization for Low Bitwidth and Hardware Compliant Neural Networks}


\author{Dominika Przewlocka-Rus}
\affiliation{%
  \institution{Meta Reality Lab Research}
  \city{Redmond, WA}
  \country{USA}
  }
\affiliation{%
  \institution{AGH UST, Krakow, Poland}
  %\city{Krakow}
  %\country{Poland}
  }
\email{dominika.przewlocka@agh.edu.pl}
\authornote{The research presented in this paper was done during employment at Meta Reality Lab Research.}

\author{Syed Shakib Sarwar}
\affiliation{%
  \institution{Meta Reality Lab Research}
  \city{Redmond, WA}
  \country{USA}}
\email{shakib7@fb.com}

\author{H. Ekin Sumbul}
\affiliation{%
  \institution{Meta Reality Lab Research}
  \city{Sunnyvale, CA}
  \country{USA}}
\email{ekinsumbul@fb.com}

\author{Yuecheng Li}
\affiliation{%
  \institution{Meta Reality Lab Research}
  \city{Pittsburgh, PA}
  \country{USA}}
\email{yuecheng.li@fb.com}

\author{Barbara De Salvo}
\affiliation{%
  \institution{Meta Reality Lab Research}
  \city{Burlingame, CA}
  \country{USA}}
\email{barbarads@fb.com}

\renewcommand{\shortauthors}{D. Przewlocka-Rus, et al.}

\begin{abstract}
  Deploying Deep Neural Networks in low-power embedded devices for real time-constrained applications requires optimization of memory and computational complexity of the networks, usually by quantizing the weights. Most of the existing works employ linear quantization which causes considerable degradation in accuracy for weight bit widths lower than 8. Since the distribution of weights is usually non-uniform (with most weights concentrated around zero), other methods, such as logarithmic quantization, are more suitable as they are able to preserve the shape of the weight distribution more precise. Moreover, using base-2 logarithmic representation allows optimizing the multiplication by replacing it with bit shifting. In this paper, we explore non-linear quantization techniques for exploiting lower bit precision and identify favorable hardware implementation options. We developed the Quantization Aware Training (QAT) algorithm that allowed training of low bit width Power-of-Two (PoT) networks and achieved accuracies on par with state-of-the-art floating point models for different tasks. We explored PoT weight encoding techniques and investigated hardware designs of MAC units for three different quantization schemes - uniform, PoT and Additive-PoT (APoT) - to show the increased efficiency when using the proposed approach. Eventually, the experiments showed that for low bit width precision, non-uniform quantization performs better than uniform, and at the same time, PoT quantization vastly reduces the computational complexity of the neural network.
\end{abstract}

\keywords{non uniform quantization, logarithmic quantization, neural networks, hardware design}

\maketitle

\section{Introduction} \label{sec:intro}
\begin{figure}[h] 
  \centering
  \includegraphics[width=1\linewidth]{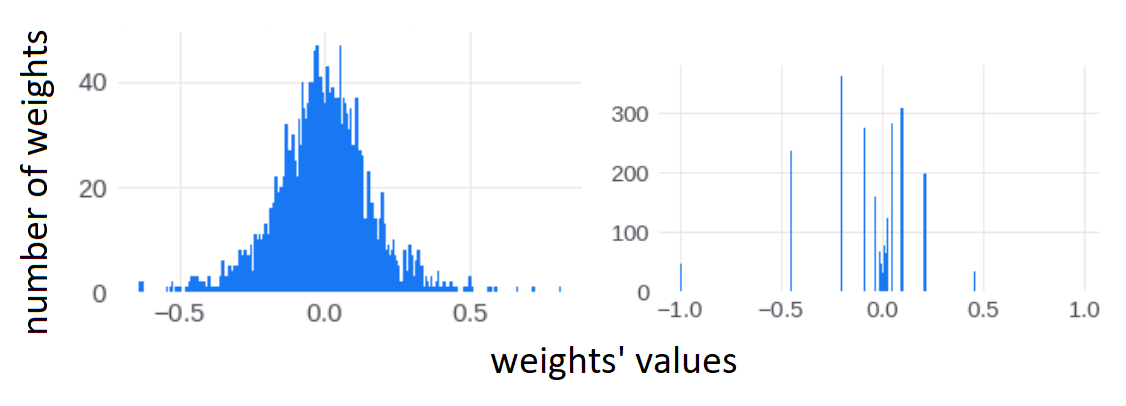}
  \caption{Log2 based (right) quantization for exemplar layer floating point weights (left).}
%   \caption{Comparison of uniform (right plot) and log2 based (bottom) quantization for exemplar layer floating point weights (left).}
  \label{weights_distribution}
\end{figure}

Image and point cloud processing algorithms are the key components of the most advanced commercial applications like AR/VR (Augmented and Virtual Reality) or autonomous vehicles (\textit{e.g.} self-driving cars, UAV). State-of-the art solutions are usually based on Deep Neural Networks (DNNs). However, in order to satisfy the low-power and real time processing requirements, various optimization techniques have to be applied to deploy the networks on embedded devices. One idea is to use lower precision arithmetic with weights and/or activations represented using 8 or less bits. The most popular uniform quantization yields near floating point accuracy using 8 bits representation. However, further precision reduction causes substantial and often unacceptable degradation. Since the distribution of the network’s weights is non-uniform - concentrated around zero and sparse away - other quantization methods, which follow the distribution more closely, should be considered. One option is to use logarithmic quantization, with increased resolution around zero (Fig. \ref{weights_distribution}). Moreover, using the logarithm base 2 (Powers-of-Two weights), we can easily use bit shifting based multiplication which vastly reduces the complexity and thus the power consumption for convolution operations in DNNs.
In this paper, we explore Powers-of-Two (PoT) weights for low precision networks and demonstrate the achievable hardware benefits using the proposed method. The main contributions are:
\begin{itemize}
    \item Proposition of two Quantization Aware Training (QAT) methods for PoT quantization and investigation of their performance.
    \item Hardware design, synthesis and analysis of three digital multipliers for uniform, PoT and Additive-Powers-of-Two (APoT) quantization for FPGA and ASIC, demonstrating the lowest resources and power consumption when using PoT weights. 
    \item Demonstration of potential memory savings  when using very low precision resulting in size compression.
\end{itemize}
To the authors’ best knowledge, this is the first paper with analysis and comparison of hardware designs of MAC units for different uniform and nonuniform quantization schemes. At the same time, using the proposed QAT algorithm, with PoT weights we achieve accuracies on par with the state-of-the-art floating point models.
\section{Related work} \label{sec:related_work}
A recent survey \cite{2.1} on efficient neural network inference summarizes different aspects of quantization, including uniform and nonuniform quantization, symmetric or asymmetric (for tailed weights’ distributions), different granularity (quantizing channel wise, kernel wise, etc.), as well as Quantization Aware Training (QAT) or Post Training Static Quantization (PTSQ). Logarithmic quantization was first introduced in \cite{2.2}. The authors performed PTSQ experiments quantizing separately activations, convolution and fully connected layers for VGG16 and AlexNet, and showed little degradation in accuracy for low bit width precision. They also proposed training with logarithmic representation. In \cite{2.3}, to increase the resolution of quantized weights around zero, the authors proposed using additional quantization levels, as fractional exponents of 2. The reported PTSQ accuracy results were better than using the algorithm proposed in \cite{2.2}. However, introducing fractional weights makes it difficult to use bit shifting instead of multiplication. Similar ideas motivated the authors of \cite{2.4}, but led to a far more complex and effective QAT algorithm for Additive-Powers-of-Two (APoT) networks, where the weights are represented by both PoT and sums of PoT values. The reported results are state-of-the-art for low bit width precision networks, preserving or even surpassing the baseline floating point accuracies for various residual networks. In the proposed method both weights and activations are quantized, excluding first and last layers. In \cite{2.5}, the authors presented two approaches to train PoT (DeepShift) networks: by learning values of i) shifts and ii) weights. For both scenarios, all convolution and fully connected layers’ multiplications are replaced with bit shifting. The reported results are close to floating point accuracies for 4 or 5 bit width precision. The authors also implemented GPU kernels and showed decreased inference time when using bit shifting instead of multiplication. Other papers showed more hardware-oriented approaches: in \cite{2.6}, the authors proposed the hardware-inspired ShiftAddNet, with layers based solely on bit shifting and add operations with an efficient training algorithm. In \cite{2.7}, an energy efficient inference engine using only bitshift-add convolutions - LogNet - was presented. The authors showed that in comparison to high-end GPUs, the proposed engine allows to significantly decrease the energy consumption per input image. 
The above-mentioned works inspired both the study of QAT for PoT networks and the investigation and design of different MAC units for hardware efficient neural networks. While previous works focused on GPU kernels or FPGA-based engines, we focused on edge deployment using ASIC designs.
\section{Logarithmic quantization} \label{sec:log_quant}
This section summarizes both the theory of logarithmic quantization and the proposed method. 
\subsection{Fundamentals} \label{sec:log_quant:subsec:basics}
Logarithmic quantization is defined with Eq. \ref{eq:loq_quant_1} and \ref{eq:loq_quant_2}, where bitwidth is the number of bits for the absolute value of weight \textit{x} (excluding the sign bit) and FSR is the Full Scale Range which determines the maximum possible quantization level value.
\begin{small}
\begin{equation} \label{eq:loq_quant_1}
  \mbox{LogQuant(x, bitwidth, FSR)} = \begin{cases} 0, & \mbox{if } x = 0 \\ 2^{\tilde{x}}, & \mbox{otherwise} \end{cases}
\end{equation}
\begin{equation} \label{eq:loq_quant_2}
 \begin{aligned}
  \tilde{x} = \mbox{Clip}(\mbox{Round}(log_{2}(|x|)), \mbox{FSR} - 2^{\mbox{bitwidth}}, \mbox{FSR}), \\
  \mbox{Clip(x, min, max)} = \begin{cases} 0, & x \leq \mbox{min} \\ \mbox{max} - 1, & x \geq \mbox{max} \\ x, & \mbox{otherwise} \end{cases}
 \end{aligned}
\end{equation}
\end{small}
Full Scale Range can be set to a fixed value or adjusted dynamically during training, given the actual distribution of weights in each epoch. FSR can be also used to discard outlying weights. It is important to note that quantization is performed for absolute values of weights and the sign is stored separately. In our experiments, before quantization we scale weights to $[-1, 1]$ range, so the value of FSR is constant (and equal to $2^0=1$), but since the scaling factor is changing based on floating point distribution in each epoch, we’re actually learning the FSR (thus we use dynamic FSR). Given the layers’ weights W, we calculate the scaling factor SF = max(abs(W)) and normalize the weights $W_N = W/SF$. Then we apply the equations \ref{eq:loq_quant_1} and \ref{eq:loq_quant_2} to obtain the quantized weights $W_Q = Q(W_N) * SF$. The quantized values $Q(W_N)$ are powers-of-two and can be represented using only exponents, with an extra bit for the weight’s sign - in the rest of the paper, we use the notation where $x$ bit width quantization means $x-1$ bits for absolute value of weight, and one bit for its sign. Since we use PoT quantization, all convolution and fully connected multiplications can be replaced with bit shifting, using the exponent value. The scaling factor (SF) is merged with batch normalization gain and does not introduce any additional computations. It is worth noting that in comparison with uniform quantization, the nonuniform quantization does not prune any weights automatically. Using the proposed approach, all weights near zero (or equal to zero) are promoted to minimum quantization level. Therefore, to introduce pruning, we propose the Pruning Factor (PF) and set the weights to zero with relation to the minimum possible quantization level - Eq. \ref{eq:pruning_factor}. 
\begin{equation} \label{eq:pruning_factor}
\centering
  w = \begin{cases} 0, & \mbox{if } w \leq PF \\ w, & \mbox{otherwise} \end{cases}
\end{equation}
\subsection{Quantization Aware Training} \label{sec:log_quant:subsec:QAT}
\begin{figure}[h] 
  \centering
  \includegraphics[width=0.75\linewidth]{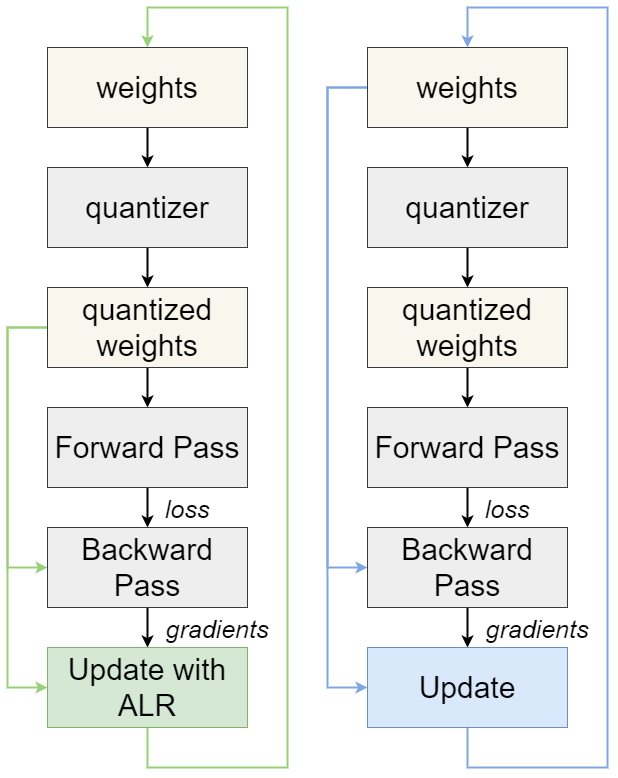}
  \caption{Flow diagrams of two training algorithms: with Adaptive Learning Rate (left) and standard Straight Through Estimator (right).}
  \label{QAT}
\end{figure}

\begin{figure}[h] 
  \centering
  \includegraphics[width=0.75\linewidth]{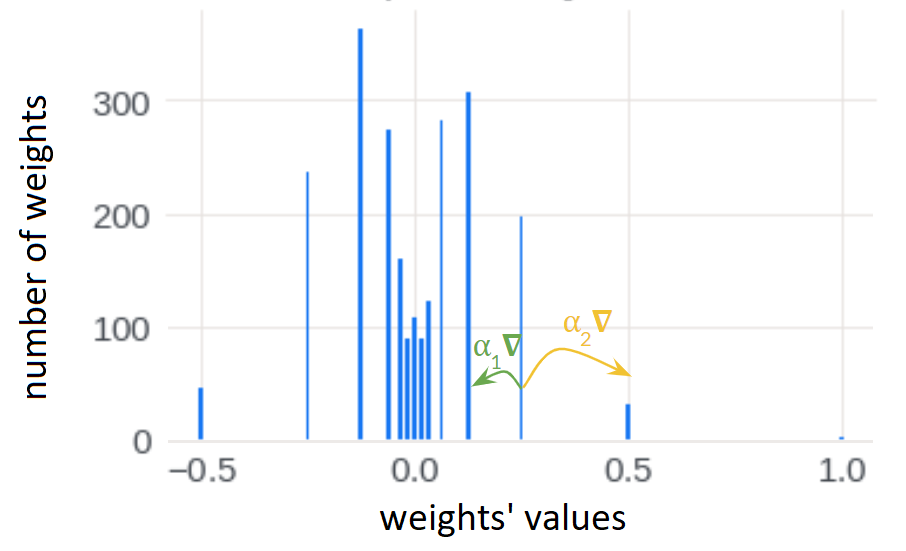}
  \caption{Illustration of Adaptive Learning Rate: to compensate for unequal distances between quantization levels we modify the base learning rate to have greater value for weights away from zero.}
  \label{ALR}
\end{figure}

We compare two approaches for Quantization Aware Training: using Adaptive Learning Rate (ALR) and Straight-Through Estimator (STE), both presented in Fig. \ref{QAT}.
In the ALR approach, we calculate both forward and backward pass using quantized weights (quantization described in Sec. \ref{sec:log_quant:subsec:basics}). It is important to note that for backward pass we do not take into account the quantization equations while calculating the gradients (we treat the quantized weights as floating point values). To compensate for unequal distances between the consecutive quantization levels (Fig. \ref{ALR}), we propose to adapt the basic learning rate with relation to current quantized weight.
The second QAT algorithm is based on a simple STE approach: in opposition to ALR, we calculate gradients using floating point values - then we update the floating point network and do quantization. This allows for a smoother transition between consecutive quantization levels. Moreover, since for low precision weights, training from scratch can lead to non-convergence, we use pre-trained floating point models for initial weights' values. Table \ref{tab:alg_comp} summarizes the differences between ours and other QAT methods from literature.
\begin{small}
\begin{table}
  \caption{Comparison of QAT algorithms}
  \label{tab:alg_comp}
  \begin{tabular}{lccc}
    \toprule
    Algorithm & Method & FSR & \makecell{Weight \\ normalization}\\
    \midrule
    Ours ALR & ALR & Dynamic & Yes \\
    Ours STE & STE & Dynamic & Yes \\
    DeepShift \cite{2.5} & STE & Fixed & No \\
    APoT \cite{2.4} & STE & SGD optimized & Yes \\
  \bottomrule
\end{tabular}
\end{table}
\end{small}
\section{Benchmark results} \label{sec:alg_results}
We tested our QAT algorithms on three datasets: CIFAR10, CIFAR100 and ImageNet. Each network is first initialized with the pre-trained floating point weights. We train with SGD for 15 epochs, with a multi step base learning rate starting from 0.001 and decreasing by 0.1 every 5 QAT epochs. We use layer wise quantization and quantize all layers - unless stated otherwise.
\subsection{CIFAR 10 and CIFAR 100} \label{sec:alg_results:subsec:cifars}
Table \ref{tab:deepshift_cifar10} summarizes the results for ResNet18 architecture trained on CIFAR10 and compares the accuracies with DeepShift \cite{2.5} (all layers are quantized with PoT weights). For both training approaches (ALR and STE) and bit width 5 and 4, the accuracy of the quantized network is close to the floating point version -- the gap increases if we further reduce the number of bits. 
\begin{table}
  \caption{Comparison with DeepShift \cite{2.5} for ResNet18 trained on CIFAR-10}
  \label{tab:deepshift_cifar10}
  \begin{tabular}{lccc}
    \toprule
    Bit width & ALR & STE & DeepShift\\
    \midrule
    Baseline & 94.14\% & 94.14\% & 94.45\% \\[3pt]
    5 & \makecell{93.79\% \\ \small{(-0.35)}} & \makecell{93.93\% \\ \small{(-0.21)}} & \makecell{94.43\% \\ \small{(-0.02)}} \\[10pt]
    4 & \makecell{93.81\% \\ \small{(-0.33)}} & \makecell{93.97\% \\ \small{(-0.17)}} & \makecell{94.38\% \\ \small{(-0.07)}} \\[10pt]
    3 & \makecell{92.94\% \\ \small{(-1.2)}} & \makecell{92.82\% \\ \small{(-1.32)}} & \makecell{92.04\% \\ \small{(-2.41)}} \\
  \bottomrule
\end{tabular}
\end{table}
Similar experiments were performed for comparison with APoT \cite{2.4} - Table \ref{tab:apot_cifar10} shows results for ResNet20 trained on CIFAR 10. It is important to note that the APoT approach 
quantizes both weights and activations. However, in APoT, first convolution and last fully connected layers were not quantized to minimize accuracy degradation due to quantization error. Hence, for closest comparison, in this set of experiments, we also quantize only hidden convolution layers. However, we did not apply quantization on activations in this work.
\begin{table}
  \caption{Comparison with APoT \cite{2.4} for ResNet20 trained on CIFAR 10}
  \label{tab:apot_cifar10}
  \begin{tabular}{lccc}
    \toprule
    Bit width & ALR & STE & APoT\\
    \midrule
    Baseline & 91.77\% & 91.77\% & 91.60\%\\[3pt]
    4 & \makecell{91.37\% \\ \small{(-0.4)}} & \makecell{91.7\% \\ \small{(-0.07)}} & \makecell{92.3\% \\ \small{(+0.7)}}\\[10pt]
    3 & \makecell{90.37\% \\ \small{(-1.4)}} & \makecell{90.88\% \\ \small{(-0.89)}} & \makecell{92.2\% \\ \small{(+0.6)}}\\
  \bottomrule
\end{tabular}
\end{table}
Both Tables \ref{tab:deepshift_cifar10} and \ref{tab:apot_cifar10} present results for highly redundant networks. We observe that the proposed solutions achieve accuracies on par with DeepShift while slightly worse than APoT. This is expected since for APoT, the resolution of the weights is greater and thus the quantization error is much less.
However, since over-parameterized networks are resilient to quantization, we performed experiments with more complex problems - CIFAR 100 and ImageNet (presented in Sec. \ref{sec:alg_results:subsec:imagenet}). From Table \ref{tab:mixed} -- with ResNet20 and CIFAR 10 and 100 results -- we can draw several conclusions.
% \begin{itemize}
%     \item 
    Firstly, for CIFAR 100 and PoT quantization of all weights, the ALR based QAT performs worse than STE (the difference in accuracy is around 4\%). 
    %This tendency will be also noticeable for the ImageNet dataset, so 
    It is reasonable to state that STE is more suitable than ALR for complex problems. 
    % \item 
    Secondly, since the last fully connected layer is quite sensitive, and it is a common practice not to quantize it so firmly, we tested a mixed quantization approach. For both CIFAR 10 and CIFAR 100, we quantize convolution layers using 4 bit width PoT weights, and the fully connected layer with 8 bit width uniform quantization. This allowed us to achieve floating point accuracy for both datasets with STE QAT. 
    % \item 
    It is also very important to acknowledge that for higher bit widths (\textit{e.g.} 8), the PoT quantization in the proposed form will perform worse than uniform quantization - increasing the bit width would only cause more tiny weights to be concentrated around 0. Thus, in this scenario, for fully connected layers we used uniform quantization.
% \end{itemize}
%
\begin{table}
  \caption{Mixed quantization experiments for ResNet20 and CIFAR datasets: convolution layers are quantized with logarithmic weights (L), while the fully connected layers are either logarithmic (L) or uniform (U)}
  \label{tab:mixed}
  \begin{tabular}{lcccc}
    \toprule
    Dataset & \makecell{Precision \\ (C/FC)} & ALR & STE & Baseline \\
    \midrule
    CIFAR 10 & 4L/4L & 91.05\% & 91.67\% & 91.77\% \\
    CIFAR 10 & 4L/8U & 91.01\% & 91.60\% & 91.77\% \\
    CIFAR 100 & 4L/4L & 63.79\% & 67.68\% & 68.68\% \\
    CIFAR 100 & 4L/8U & 66.19\% & 68.51\% & 68.68\% \\
  \bottomrule
\end{tabular}
\end{table}
\subsection{ImageNet} \label{sec:alg_results:subsec:imagenet}
In Table \ref{tab:resnets_imagenet}, we summarized the ImageNet experiments for residual networks. We achieve near floating point accuracy for PoT networks. At the same time our results are comparative (usually slightly better) to the state-of-the-art solutions - both DeepShift and APoT. For STE QAT we see that there is little degradation or even improvement in accuracy (for more redundant networks) in comparison to floating point networks. For similar results with ALR the mixed quantization had to be used -- STE QAT performs better, like in previous CIFAR  experiments (Table \ref{tab:mixed}).
\begin{small}
\begin{table}
  \caption{Results for ImageNet dataset with residual networks’ architectures. Within the parenthesis, we show the differences with the baseline floating point networks for each experiment (the baseline accuracies are different for ours, DeepShift and APoT experiments)}
  \label{tab:resnets_imagenet}
  \begin{tabular}{lccccc}
    \toprule
    Network & \makecell{Precision \\ (C/FC)} & ALR & STE & DeepShift & APoT*\\
    \midrule
    ResNet18 & 4L/32F & \makecell{68.774\% \\ \small{(-0.984)}} & \makecell{69.982\% \\ \small{(+0.224)}} & - & \makecell{70.7\% \\ \small{(+0.5)}}\\[10pt]
    ResNet18 & 4L/8U & \makecell{68.56\% \\ \small{(-1)}} & \makecell{69.868\% \\ \small{(+0.11)}} & - & -\\[10pt]
    ResNet18 & 4L/4L & \makecell{67.076\% \\ \small{(-2.682)}} & \makecell{69.526\% \\ \small{(-0.232)}} & \makecell{69.56\% \\ \small{(-0.198)}} & -\\[10pt]
    ResNet50 & 5L/5L & \makecell{75.14\% \\ \small{(-0.99\%)}} & \makecell{76.468\% \\ \small{(+0.338)}} & \makecell{76.33\% \\ \small{(+0.216)}}\\[10pt]
    ResNet50 & 4L/32F & \makecell{75.642\% \\ \small{(-0.488)}} & \makecell{76.404\% \\ \small{(+0.274)}} & - & \makecell{76.6\% \\ \small{(+0.2)}}\\[10pt]
    ResNet50 & 4L/8U & \makecell{75.582\% \\ \small{(-0.548)}} & \makecell{76.384\% \\ \small{(+0.254)}} & - & -\\[10pt]
    ResNet50 & 4L/4L & \makecell{75.086\% \\ \small{(-1.044)}} & \makecell{76.314\% \\ \small{(+0.184)}} & - & -\\
  \bottomrule
  \multicolumn{6}{l}{\small *APoT does not quantize first layer but does quantize activation} \\
\end{tabular}
\end{table}
\end{small}
Moreover, we also tested the proposed approach with other non-residual architectures (Table \ref{tab:nonresidual_imagenet}). For VGG-11 (4 bit width) and VGG-16 (5 bit width) we are on par with floating point baselines. For MobileNet V2 we observe a 3.5\% degradation in accuracy - however, it is important to note that MobileNet has a very compact architecture compared to other tested networks. Hence, such a gap between low precision and floating point models is understandable. The presented results are only for the STE QAT algorithm since the ALR did not perform well for non residual networks. It is also worth noting that very few experiments on networks different than ResNets and less redundant networks with PoT quantization are reported in literature.
\begin{small}
\begin{table}
  \caption{Accuracies for low precision non residual architectures and ImageNet dataset}
  \label{tab:nonresidual_imagenet}
  \begin{tabular}{lcccc}
    \toprule
    Dataset & \makecell{Precision \\ (Conv/FC)} & STE & DeepShift & Baseline \\
    \midrule
    MobileNet V2 & 4L/4L & \makecell{68.376\% \\ \small{(-3.502)}} & - & 71.878\%\\[10pt]
    \makecell[l]{VGG-11 \\ (with BN)} & 4L/4L & \makecell{70.326\% \\ \small{(-0.044)}} & - & 70.370\%\\[10pt]
    VGG-16 & 5L/5L & \makecell{71.432\% \\ \small{(-0.16)}} & \makecell{71.56\% \\ \small{(-0.03)}} & 71.592\%\\[10pt]
    \makecell[l]{VGG-16 \\ (with BN)} & 5L/5L & \makecell{73.77\% \\ \small{(+0.41)}} & - & 73.360\%\\
  \bottomrule
\end{tabular}
\end{table}
\end{small}
\subsection{Comparison with Uniform Quantization} \label{sec:alg_results:subsec:uniform}
It is also essential to show that for low bit width precision, the logarithmic (PoT) quantization actually is performing better than uniform quantization. Table \ref{tab:uniform_vs_nonuniform} compares results for three different datasets and both uniformly and non-uniformly quantized networks. While for simple problems and redundant networks the gap between PoT and uniform quantization is not high, for ResNet18 and ImageNet the difference in accuracy is significant (around 10\%). This shows that for low bit width precision, to preserve near floating point accuracy in complex tasks, using PoT quantization is advisable. 
\begin{small}
\begin{table}
  \caption{Comparison between PoT and uniform quantization. Only convolution layers are quantized to 4 bit width precision}
  \label{tab:uniform_vs_nonuniform}
  \begin{tabular}{lcccc}
    \toprule
    Network & Baseline & ALR & STE & Uniform\\
    \midrule
    ResNet20 CIFAR 10 & 91.77\% & \makecell{91.05\% \\ \small{(-0.72)}} & \makecell{91.6\% \\ \small{(-0.17)}} & \makecell{91.22\% \\ \small{(-0.55)}}\\[10pt]
    ResNet20 CIFAR 100 & 68.68\% & \makecell{66.27\% \\ \small{(-2.41)}} & \makecell{68.51\% \\ \small{(-0.17)}} & \makecell{65.47\% \\ \small{(-3.21)}}\\[10pt]
    ResNet18 ImageNet & 69.76\% & \makecell{68.77\% \\ \small{(-0.99)}} & \makecell{69.868\% \\ \small{(+0.11)}} & \makecell{57.83\% \\ \small{(-10.93)}}\\
  \bottomrule
\end{tabular}
\end{table}
\end{small}
\subsection{Pruning} \label{sec:alg_results:subsec:Pruning}
Since logarithmic quantization does not automatically prune weights (in contrast to uniform quantization when applied for bit precision much lower than 8), we decided to run additional tests with the Pruning Factor defined in Eq. \ref{eq:pruning_factor}. Table \ref{tab:pruning} shows the results for 4 bit width ResNet20 trained on CIFAR 100. We see that even if we skip the smallest quantization level (with PF=2), we set to zero no more than 11\% weights. Pruning can firmly increase the memory savings (both in terms of area as well as number of memory accesses). Hence, it would be advisable to seek for other methods to increase the sparsity of the network. For example, instead of discarding lowest quantization levels, we should try to move them further from zero. This would allow the use of higher threshold values, but would not result in decreasing the number of quantization levels.
\begin{table}
  \caption{Pruning experiments for ResNet20 CIFAR 100 and 4 bit width logarithmic quantization of all layers. Baseline accuracy: 68.68\%}
  \label{tab:pruning}
  \begin{tabular}{ccc}
    \toprule
    Pruning Factor & Pruned weights & Accuracy\\
    \midrule
    2 & 10.9\% & 67.39\%\\
    1 & 5.9\% & 67.49\%\\
    0.5 & 3.4\% & 67.49\%\\
    0 & 0\% & 67.68\%\\
  \bottomrule
\end{tabular}
\end{table}
\section{Hardware Benefits} \label{sec:hw}
The obvious benefit from using PoT weights is replacing the standard multipliers in MAC (Multiply-And-Accumulate) Units with bit shifting. The introduced scaling factor is merged with batch normalization gain and the weights are stored as exponent values with sign (bit shift values with shift direction), so no additional memory or decoding logic is needed. The potential hardware benefits are considered for both existing off-shelf devices, as well as FPGAs or ASICs.
\subsection{Memory Cost} \label{sec:hw:subsec:memory}
In hardware, the memory cost consists of two factors: number of bits for storing the weights (memory capacity) and number of memory transactions.
Quantizing weights to 4 bits vastly reduces the required memory capacity compared to 8 bit quantized weights. Furthermore, low memory capacity requirement also leads to lower leakage power consumption.
On the other hand, by packing more weights into a single memory word, 4 bit quantization further reduces the number of memory transactions, which lowers the overall memory access energy cost.
To evaluate it, we decided to use Ethos-u Vela compiler \cite{ethosuvela} targeting Ethos-U55 micro neural processing unit \cite{ethosu55} on our trained tflite models. We generated the tflite models from PyTorch using PyTorch-ONNX-tflite flow to generate required INT8 quantized network format for deploying in the ARM NPU. That is, our 4 bit width networks were assumed to have 8 bit width, but the values of weights remain unchanged. 
While preparing the model for deployment, the Vela compiler compresses the weights based on actual number of bits required to represent the weight values and amount of sparsity (number of zeros) present in the distribution. 
Hence, it is able to utilize only 4 bits per weight for our trained models when placing weights in memory. 
The results for ResNet20 CIFAR 100 are presented in Table \ref{tab:hw_memory}. First row shows the results for a floating point network converted to INT8 using Tensorflow to tflite conversion. We see that there is minimal (<10\%) weights’ compression -- so the size of the Vela generated model is close to the size of the tflite INT8 model (which is 317 KB). The second row shows results for the uniformly quantized network, with 41\% of weights pruned to 0 (with automatic pruning during quantization). The size of the model decreased from 265KB for INT8 to 138KB. The results for PoT networks and different levels of pruning show that the size of the Vela generated network is slightly larger compared to the auto pruned network in row 2. This results from the higher sparsity of the uniformly quantized network and demonstrates the positive impact of pruning for hardware design. However, even for the not so sparse PoT networks, the compression is significant - and the accuracy of the network is higher than in case of uniform quantization. The accuracy difference is more significant for more complex tasks such as ImageNet classification (Tab. \ref{tab:uniform_vs_nonuniform}).
For custom designs, we should also expect the reduction of the number of needed memory access (for example, by roughly 2x if we switch from INT8 to INT4 weights) which would translate into the decreased number of memory operations per second (thus reducing the inference time and energy).
Moreover, the lower precision weights will lead to reduction in memory bandwidth requirement (as we can pack more weights for a~single transfer).
\begin{small}
\begin{table}
  \caption{Memory savings (model size compression) for 4 bit width quantized neural networks using Ethos-u Vela compiler. For this experiment ResNet20 CIFAR 100 was used, the original model size is 317 KB}
  \label{tab:hw_memory}
  \begin{tabular}{lccc}
    \toprule
    Features & \makecell{Weights \\ compression} & Size KB & Accuracy\\
    \midrule
    Baseline INT8 & 0.93 & 268 & 68.68\%\\
    4U/32F  (41\% of zeros) & 0.45 & 138 & 65.47\%\\
    4L/4L (no pruning) & 0.49 & 150 & 67.68\%\\
    4L/4L (8\% of zeros) & 0.50 & 152 & 67.49\%\\
    4L/4L (11\% of zeros) & 0.48 & 149 & 67.39\%\\
  \bottomrule
\end{tabular}
\end{table}
\end{small}
\subsection{Shift-based Artificial Neuron} \label{sec:hw:subsec:MAC}
Finally, we compare the hardware designs of the MAC units for uniform and non-uniform quantization. The baseline for comparison is a standard uniform multiplier with 8 bit input operands (both input activation and weight have 8 bit precision), which is common in off-the-shelf edge HW accelerators for DNNs. For low bit width (4 bit) weights, we consider three different MAC designs based on 4bx8b multiplication: regular multiplier for uniform quantization, and shift-based multiplier for PoT weights and for APoT weights.
Our hardware designs are based on typical off-the-shelf multiplier, accumulator and shifter designs (without any custom design optimization).
Note that we have assumed 8 bit activation for all designs.

\begin{figure}[h] 
  \centering
  \includegraphics[width=0.85\linewidth]{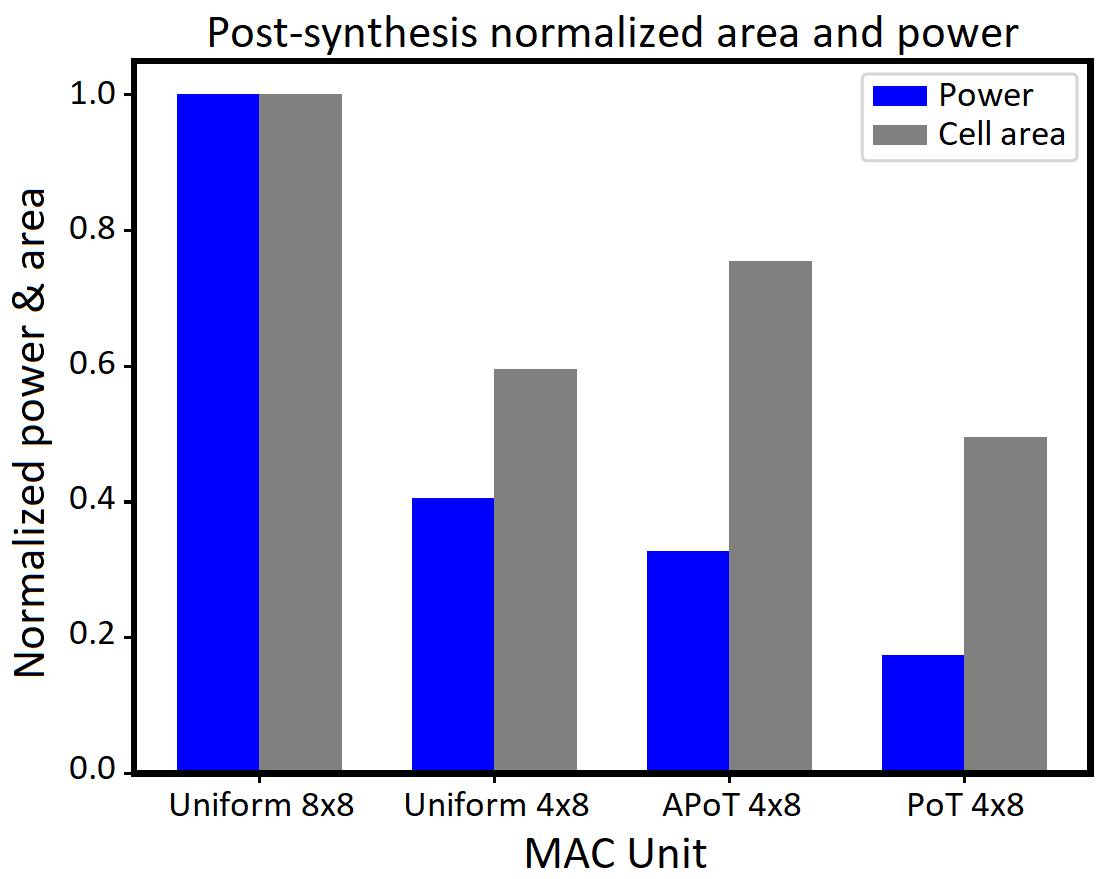}
  \caption{Comparison of post-synthesis area and power usage for Samsung 5nm.}
  \label{asic}
\end{figure}

% Since uniform quantization introduces pruning, it is reasonable to include 0 checking to reduce the number of multiplications if the weight is equal to 0 (by skipping the operation). For PoT and APoT approaches, the 0 checking can be introduced in future, when the sparsity of the log2 network is increased. 
For PoT and APoT quantization, the weight is stored separately as sign and exponent (bit shifting value). Therefore, after shifting, we need to make sign correction. It is clear that for APoT the logic is more complex and requires more resources since we need to perform the bit shifting twice and then sum both partial results. Moreover, this solution will also require even more additional logic since we need to decode the two APoT factors from the value read from the memory (this decoding is not taken into account in our designs).
Table \ref{tab:fpga} presents the synthesis results of the proposed MAC unit designs for FPGA. For proper comparison we disabled the usage of DSP.
% To demonstrate gains from reducing the weights’ bit width from 8 to 4, we synthesized both 8x8 and 4x8 multiplier-based MAC units (for proper comparison, we disabled the usage of DSP). 
Switching from 8x8 to 4x8 multiplication caused the number of used resources to decrease by roughly half for both LUTs and FFs. Further reduction, however not so firm, is observed when comparing uniform 4x8 and PoT MAC units. Although the bit shifting operation is far less complex than multiplication, the sign correction introduces additional logic. Due to the doubled number of operations, the APoT Unit uses roughly twice as much LUTs and FFs as the PoT Unit. At the same time, it is less area efficient than Uniform 4x8 MAC Unit. 
\begin{table}
  \caption{Vivado post-synthesis  resource usage for different MAC Units (for Zynq-7000 \cite{zynq7000} platform). For uniform MAC the DSP usage was disabled for proper comparison with multiplier-less unit}
  \label{tab:fpga}
  \begin{tabular}{lcc}
    \toprule
    MAC PE & LUT & FF\\
    \midrule
    Uniform 8x8 & 87 & 39\\
    Uniform 4x8 & 46 & 27\\
    APoT 4x8 & 55 & 49\\
    PoT 4x8 & 39 & 25\\
  \bottomrule
\end{tabular}
\end{table}
The final step of our analysis involves synthesizing the proposed designs with 5nm technology library from a leading foundry. This part is aiming to quantify the benefits of PoT quantization in ASIC deployment.
Proposed MAC configurations are implemented in RTL-level and synthesized to gate netlists targeting 250 MHz clock frequency (at TT corner, 0.6V, 25\degree C) while being optimized for low-power consumption during synthesis. We used 50\% toggling rate for all designs.
The power and area results with relation to the 8x8 uniform MAC unit are presented in Fig. \ref{asic}. For both power and area, we use only the combinational part of our design. Note that for 4x8 MAC, we use 12 bit intermediate value and 16 bit accumulator. For 8x8 MAC design, we used 16b intermediate value and 24b accumulator.
Using 4 bit width weight precision, we are able to decrease the energy consumption by roughly 2.5x for uniform, 3x for APoT and 6x for PoT MAC unit, compared to traditional 8x8 uniform MAC unit. At the same time, comparing to 4x8 uniform MAC unit, PoT enables power reduction by more than 2x.
The area is also reduced by about 1.7x, 1.3x and 2x for 4x8 uniform, APoT and PoT MAC unit, respectively, compared to uniform 8x8 unit. 
The obvious conclusions are: (1) with 4 bit width weights we can reduce the energy consumption and cell area by more than 50\%, regardless of the used quantization scheme (uniform and PoT); (2) using PoT and APoT weights allows for even further reduction of the compute energy by replacing the multiplication with bit shift. The introduced sign correction combined with bit shift is still far more energy efficient than multiplication; (3) using PoT and bit shifting allows reduction of power consumption by more than 2x comparing to standard low-precision multiplication.
\section{Conclusion} \label{sec:conclusion}
8 bit width uniform quantization allows to achieve state-of-the-art floating point accuracies, but at the same time is not sufficient if we target low-power, embedded devices, since the memory complexity and power consumption of 8x8 multipliers are high, as demonstrated in the proposed paper. On the other hand, using lower bit widths causes the accuracy to drop significantly for complex problems. To overcome this issue, for low bit width networks, it is advisable to use non uniform quantization, which allows preservation of the weights’ distribution shape. In this paper, we proposed a STE based QAT algorithm for Power-of-Two networks and achieved state-of-the-art, floating point accuracies quantizing all convolution and fully connected layers to low bit width precision for different problems (network architectures and datasets). Use of dynamic FSR allowed us to surpass the accuracy for ImageNet trained networks reported in DeepShift \cite{2.5}. Moreover, we demonstrated that using only PoT weights vastly reduces energy consumption of MAC units used in convolution and fully connected layers, since it allows the multipliers to be replaced by shifters. Finally, the proposed method, in comparison to the APoT approach, does not require any additional weights’ decoding method which further influences the area and energy usage. Due to reduced number of bits for weights’ storage, the number of DRAM transactions is reduced, which directly speeds up the inference. At the same time, storage capacity/area requirement is reduced in addition to reduction in memory leakage power.
In conclusion, the proposed solution not only allows floating point precision, but is well fitted for low power and real time embedded systems.
\begin{acks}
We would like to thank Pietro Caragiulo and Manzur Yazdani for their support with the hardware design synthesis tools, as well as Zhongnan Qu for his involvement in project discussions.
\end{acks}
\bibliographystyle{ACM-Reference-Format}
\bibliography{acmart}
\end{document}